\documentclass[conference]{IEEEtran}
\IEEEoverridecommandlockouts
\usepackage{stfloats}

\usepackage{cite}
\usepackage{amsmath,amssymb,amsfonts}
\usepackage{algorithmic}
\usepackage{algorithm2e}  
\usepackage{graphicx}
\usepackage{textcomp}
\usepackage{comment}
\usepackage{xcolor}
\usepackage{bm}
\usepackage{booktabs}
\usepackage{multirow}
\usepackage{hyperref}
\usepackage{float}
\newfloat{footnote}{hb}

\def\BibTeX{{\rm B\kern-.05em{\sc i\kern-.025em b}\kern-.08em
    T\kern-.1667em\lower.7ex\hbox{E}\kern-.125emX}}
\begin{document}

\title{Neuromorphic Wireless Device-Edge Co-Inference via the Directed Information Bottleneck\\

\thanks{The work of Y. Ke, Z. Utkovski, M. Heshmati, J. Dommel and S. Stanczak was supported by the Federal
Ministry of Education and Research of Germany in the program
“Souverän. Digital. Vernetzt.” Joint project 6G Research and
Innovation Cluster (6G-RIC), project identification number 
16KISK020K. 
The work of O. Simeone was partially supported by the European Union’s Horizon Europe project CENTRIC (101096379), by the Open Fellowships of the EPSRC (EP/W024101/1), by the EPSRC project (EP/X011852/1), and by Project REASON,
a UK Government funded project under the Future Open Networks Research Challenge (FONRC) sponsored by the Department of
Science Innovation and Technology (DSIT).}
}

\author{\IEEEauthorblockA{Yuzhen Ke$^{1}$, Zoran Utkovski$^{1}$, Mehdi Heshmati$^{1}$, Osvaldo Simeone$^{2}$, Johannes Dommel$^{1}$, Slawomir Stanczak$^{1}$,$^{3}$}
\IEEEauthorblockA{$^{1}$ Department of Wireless Communications and Networks, Fraunhofer Heinrich Hertz Institute, Berlin, Germany}
\IEEEauthorblockA{$^{2}$ KCLIP, CIIPS,  Department of Engineering, King’s College London, UK}
\IEEEauthorblockA{$^{3}$ Chair of Network Information Theory, Technical University of Berlin, Germany} 
}

\maketitle

\begin{abstract}
An important use case of next-generation wireless systems is device-edge co-inference, where a semantic task is partitioned between a device and an edge server. The device carries out data collection and partial processing of the data, while the remote server completes the given task based on information received from the device. It is often required that processing and communication be run as efficiently
as possible at the device, while more computing resources
are available at the edge. To address such scenarios, we introduce
a new system solution, termed neuromorphic wireless device-edge co-inference. According to it, the device runs sensing, processing,
and communication units using neuromorphic hardware, while the
server employs conventional radio and computing technologies. The proposed system is designed using a transmitter-centric information-theoretic criterion that targets a reduction of the communication overhead, while retaining the most relevant
information for the end-to-end semantic task of interest. Numerical results on standard data sets validate the proposed architecture, and a preliminary testbed realization is reported.

\end{abstract}


\section{Introduction}

\subsection{Neuromorphic Wireless Device-Edge Co-Inference}
\begin{figure}[t]
\centerline{\includegraphics[width=0.48\textwidth]{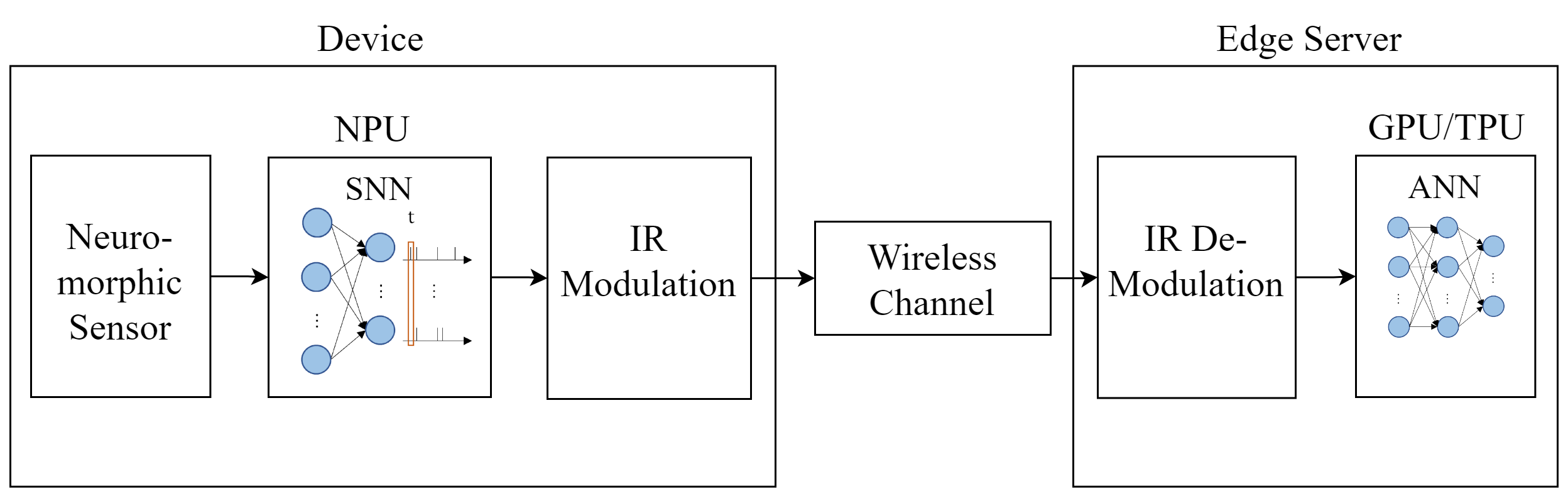}}
\caption{Neuromorphic wireless device-edge co-inference.}
\label{fig:NeuroComm_BlockDiagram}
\end{figure}

Consider the edge computing setting depicted in Fig.~\ref{fig:NeuroComm_BlockDiagram}, 
in which a device equipped with a sensor communicates over a short-range link to an edge server connected to a base station. In such a configuration, it is essential to keep the operation of the device, across processing and communication units, as efficient as possible, while accounting for the fact that more resources are typically available at the edge. In this paper, we introduce a new system solution, termed \emph{neuromorphic wireless device-edge co-inference}. In the proposed approach, the device runs sensing, processing, and communication using neuromorphic hardware, while the server carries out conventional radio and computing technologies. In this way, the edge device benefits from event-driven processing and communications, while high performance is guaranteed by leveraging resources at the edge.

As illustrated in Fig.~\ref{fig:1}, the proposed  architecture incorporates: \begin{itemize} \item \textit{a semantic  encoder} on the  device, implemented by a \emph{neuromorphic processing unit} (NPU), running a \emph{spiking neural network}  (SNN); \item  an \emph{impulse radio} (IR) wireless transmission module  performing pulse-based modulation  in an event-driven fashion to convey the output of the NPU to the edge server; and \item
a \textit{semantic decoder} implemented at the edge server in the form of an inference artificial neural network (ANN) running on a conventional accelerator such as a GPU or a TPU, which performs inference to execute the task of interest.  \end{itemize}

The proposed system is designed using an information-theoretic criterion (based on the directed information bottleneck) that targets a reduction of the communication overhead, while retaining the most relevant information for the end-to-end semantic task of interest.

\subsection{Related Work }

Neuromorphic computing is emerging as a promising technology for energy-efficient processing, particularly for semantic tasks involving time series \cite{davies2021advancing,rajendran2023towards}. The architecture proposed in this paper builds on the neuromorphic cognition system presented in \cite{skatchkovsky2020end,chen2022neuromorphic,chen2023neuromorphic}, in which a co-inference task is carried out jointly at two devices connected over a wireless channel. The two devices are equipped with NPUs, and communicate by using an IR interface. The system senses, communicates, and processes information by encoding information in the timing of spikes. The architecture introduced by  \cite{skatchkovsky2020end,chen2022neuromorphic,chen2023neuromorphic} is particularly well suited for settings in which both transmitter and receiver are energy constrained. In contrast, this paper studies an edge computing scenario in which the receiver at the edge server is not constrained by resource requirements, and thus it is not required to implement an NPU. This creates the new challenge of coordinating and jointly designing different information processing modes at the transmitter and the receiver.

References \cite{skatchkovsky2020end,chen2022neuromorphic,chen2023neuromorphic} adopted an end-to-end training of the SNNs at transmitter and receiver for a specific task under power constraints. In contrast, this paper adopts a design criterion that targets only the encoding SNN, allowing for the separate design of the receiver. This follows the general principle  followed by standardization bodies of specifying the operation of the transmitter, while leaving the receiver design open.

The proposed criterion explicitly penalizes the transmission overhead with the aim of balancing transmission resources and end-to-end performance. In this regard, the proposed design criterion may be considered as a hybrid neuromorphic-classical counterpart to the task-oriented objective in  \cite{shao2021learning}. In particular, as in \cite{shao2021learning}, we perform end-to-end learning via the information bottleneck principle, which formalizes a rate-distortion trade-off between the compression level of the encoded signals and the inference performance. 

More precisely, since, unlike \cite{shao2021learning}, we target applications involving   neuromorphic processing of time series, we adopt the \textit{directed} information bottleneck formulation introduced in \cite{skatchkovsky2021learning}. Note that in  \cite{skatchkovsky2021learning} a hybrid SNN-ANN variational autoencoder architecture was presented assuming noiseless communication between transmitter and receiver. Unlike \cite{skatchkovsky2021learning}, our study is designed for goal-oriented communication in a wireless communication setting.

\subsection{Main Contributions and Paper Organization}
In this paper, we introduce the neuromorphic wireless device-edge co-inference depicted in Fig. 1 and Fig. 2, and our main contributions are as follows.

\begin{itemize}
\item We introduce a novel hybrid neuromorphic-classical communication and computation architecture for device-edge co-inference that targets energy efficiency at the end device, while leveraging the computational power of the edge.
\item We present a novel design criterion based on the directed information bottleneck that accounts for the presence of a noisy channel between transmitter and receiver, while allowing for a controllable trade-off between communication overhead and end-to-end performance. Unlike prior work, the criterion targets exclusively the design of the encoder, leaving open the possibility to design different decoders at the edge side.
\item We validate the proposed solution using the standard MNIST-DVS and N-MNIST data sets, demonstrating significant improvements in terms of accuracy and energy efficiency over the state of the art.
\item We outline a testbed implementation involving a robot wirelessly controlled to mimic the gestures of a user captured via a remote neuromorphic camera.
\end{itemize}
\subsection{Organization}
This paper is outlined as follows. In Sec.~\ref{sec:preliminaries} we introduce the necessary information-theoretic preliminaries and provide background information on probabilistic SNNs. In Sec.~\ref{sec:system_model} we describe the wireless neuromorphic device-edge co-inference model under study. Sec.~\ref{sec:S-VDIB} introduces the encoder design framework following a reformulation of the variational directed information bottleneck principle. In Sec.~\ref{sec:results} we present numerical results and evaluate the performance of the proposed design on standard datasets. In  Sec.~\ref{sec:testbed} we describe a preliminary testbed setup to validate the proposed system solution. Sec.~\ref{sec:conclusions} concludes
the paper.


\section{Preliminaries}
\label{sec:preliminaries}
In this section, we present necessary background on SNNs and on the formalism of causal conditioning and directed information. 

\subsection{Probabilistic Spiking Neural Networks}

While the proposed architecture can be implemented with any SNN model, in this work we focus on the probabilistic SNN model~\cite{jang2019introduction} to facilitate the design via information-theoretic measures. Accordingly, we consider a set of neurons $\mathcal{V}$, where each neuron $i\in\mathcal{V}$ outputs a binary signal $s_{i,t}\in\{0,1\}$ at times $t = 0, 1, 2, \ldots$, with a value $s_{i,t} = 1$ corresponding
to a spike emitted at time $t$. Each neuron $i\in\mathcal{V}$ receives the signals emitted by a subset $\mathcal{P}_i$ of neurons through directed links (synapses). Neurons in the set $\mathcal{P}_i$ are referred to as presynaptic for postsynaptic neuron $i$. Further, we denote $\bm{s}_{i, \leq t} = (s_{i,0}, \ldots, s_{i,t})$ as the spiking history of neuron $i$, i.e. the signal emitted by neuron $i$ up to time $t$. 

In the probabilistic model, the internal state of each spiking neuron $i\in \mathcal{V}$ is defined by its membrane potential $u_{i,t}$, whose value indicates the probability of neuron $i$ to spike. The membrane potential depends on the outputs of the presynaptic neurons and on the past spiking behavior of the neuron itself, where both contributions are filtered by the feedforward kernel $\bm{a}_{t}$ and the feedback kernel $\bm{b_}t$, accounting for the synaptic delay and for the refractory period, respectively. 

The dynamic evolution of the membrane potential $u_{i,t}$ is modeled as \cite{jang2019introduction}
 \begin{equation}
 u_{i,t} = \sum_{j \in \mathcal{P}_i} w_{j,i}\left ( \bm{a}_{t} \ast \bm{s}_{j, \leq t}\right ) + w_{i}\left ( \bm{b}_{t} \ast \bm{s}_{i, \leq t-1}\right ) + \gamma_{i},
\label{eq:membrane_potential} 
\end{equation}
where $w_{j,i}$ is the synaptic weight from neuron $j$ to $i$, $w_{i}$ is a feedback weight, and $\gamma_{i}$ is a bias parameter. In (\ref{eq:membrane_potential}), $*$ denotes the convolution operator, i.e.,  $\bm{f}_t*\bm{g}_t=\sum_{\delta\geq 0} f_{\delta} g_{t-\delta}$. 

The probability of neuron $i$ emitting a spike at time $t$ is determined by the membrane potential $u_{i,t}$ as \cite{jang2019introduction}
 \begin{equation}
     p \left ( s_{i,t} = 1 |u_{i,t} \right ) = \sigma (u_{i,t}),  
     \label{MP}
\end{equation}
where $\sigma(\cdot)$ is the sigmoid function. Given the spiking probability $\sigma (u_{i,t})$, the output $s_{i,t}$ is distributed according to the Bernoulli distribution,  $s_{i,t}\sim \textrm{Bern}\left ( \sigma (u_{i,t}) \right )$. Hence, the log probability of the spike signals emitted by neuron $i$ at time $t$ can be written as  
\begin{align}
    \log p\left ( s_{i,t}\mid u_{i,t} \right ) &= \log \sigma (u_{i,t})^{s_{i,t}}\left ( 1-\sigma (u_{i,t}) \right )^{1-s_{i,t}} \\
    &= s_{i,t} \log \sigma (u_{i,t}) + \left ( 1 - s_{i,t} \right ) \log \left ( 1- \sigma (u_{i,t}) \right ).
    \nonumber
\end{align}

\subsection{Causal Conditioning and Directed Information }
For two jointly distributed random vectors (sequences) ${\bm{X}}^T = \left\{X_{1}, \ldots, X_{T}\right\}$ and $\bm{Y}^T= \left\{Y_{1}, \ldots, Y_{T}\right\}$ from time $1$ to $T$, the \emph{causally conditioned distribution} of $\bm{X}^T$ given $\bm{Y}^T$ is defined as \cite{kramer1998directed}
 \begin{equation}\label{eq:cc}
p\left (\bm{x}^T\parallel \bm{y}^T\right):=\prod_{t=1}^{T} p \left ( x_{t} \right | \bm{x}^{t-1}, \bm{y}^{t}),  
\end{equation}
where we have used $\bm{x}^t=\left\{x_1, \ldots, x_t\right\}$ to denote the realization of the random vector $\bm{X}^t=\left\{X_{1}, \ldots, X_{t}\right\}$. 

Following \cite{skatchkovsky2021learning}, we use the directed information to quantify the information flow, i.e., the causal statistical dependence, from $\bm{X}^T$ to $\bm{Y}^T$. The \emph{directed information} from  $\bm{X}^T$ to $\bm{Y}^T$ is defined as \cite{kramer1998directed}
\begin{align}
I \left ( \bm{X}^T \to \bm{Y}^T \right ) &:= H\left ( \bm{Y}^{T} \right ) - H\left ( \bm{Y}^{T} \parallel \bm{X}^{T}\right ) \nonumber \\
&:= \sum_{t=1}^{T} I\left ( \bm{X}^{t}; Y_{t}|\bm{Y}^{t-1}\right ),
\end{align}
where $I\left ( \bm{X}^{t}; Y_{t}|\bm{Y}^{t-1}\right )$ is the conditional mutual information between $\bm{X}^{t}$ and $ Y_{t}$ given $\bm{Y}^{t-1}$. Unless stated otherwise, for brevity, in the following we will omit the superscript $^T$ and will write $\bm{X}$ and $\bm{x}$ to denote the random vector $\bm{X}^T$, respectively its realization $\bm{x}^T$. 

\section{System Model}
\label{sec:system_model}

\begin{figure*}[t]
\centerline{\includegraphics[width=0.7\textwidth]{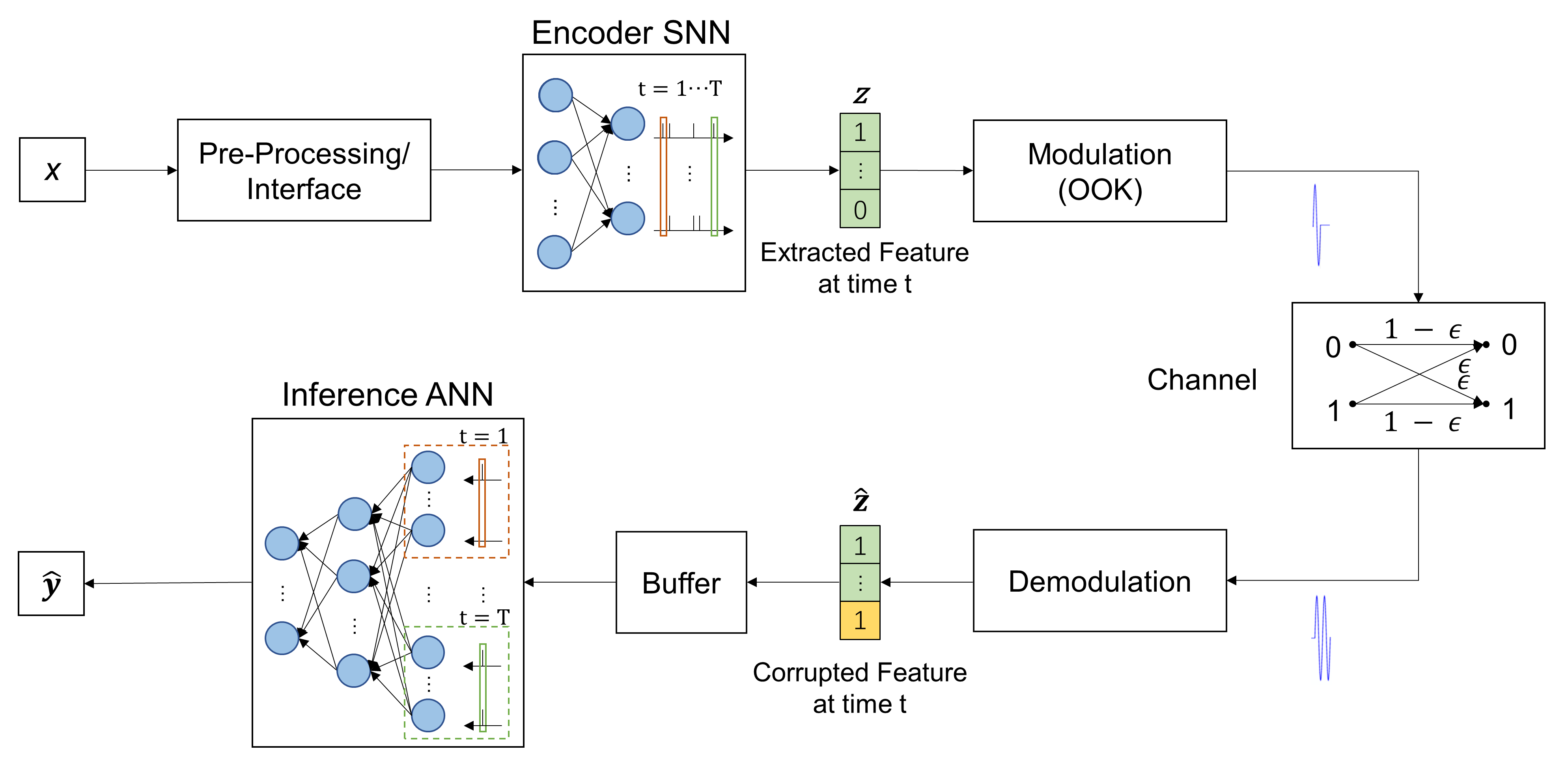}}
\caption{The proposed hybrid neuromorphic-classical  wireless device-edge co-inference solution encompassing an encoding SNN at the transmitter and a decoding ANN at the receiver with IR-based transmission.}
\label{fig:1}
\end{figure*}

We consider the device-edge co-inference system illustrated in Fig.~\ref{fig:1}. In this architecture, the device and the edge server cooperate to perform a semantic task on signals sensed by the device, with the final output produced by the edge. The system is comprised of three modules: an on-device SNN encoder, a wireless channel, and a conventional inference network implemented on the edge server. The input
data $\bm{x}$, sensed by the device, is associated with a target $\bm{y}$, which depends on the specific semantic task. For instance, it may be a single label in a classification or regression task, or a vector label for a segmentation task. Input and target output are assumed to be realizations of a pair of random variables (vectors) $(\bm{X}, \bm{Y})$. 

\subsection{Transmitter}

At the transmitter side, processing is implemented by an NPU running  a probabilistic SNN model parameterized by a vector $\bm{\phi}$. The SNN has $k$ read-out neurons producing signals  $\bm{z}$. By \eqref{MP}, the probability of the $k$-dimensional binary vector sequence $\bm{z}$ conditioned on the input $\bm{x}$ is given by 
\begin{align}
p_{\bm{\phi}}\left (\bm{z}\parallel \bm{x}  \right ) &=\prod_{t=1}^{T} p \left ( \bm{z}_{t} | \bm{z}^{t-1},\bm{x}^{t} \right ) \nonumber \\  
&= \prod_{t=1}^{T} \prod_{i \in \mathcal{V}} p_{\bm{\phi}} \left (  z_{i,t}  |  u_{i,t} \right ),
\label{eq:SNN_model}
\end{align}
where we recall that 
$u_{i,t}$ is the membrane potential of the $i$-th neuron at time $t$.

The transmitter communicates using an IR module. Accordingly,  at each time instant $t$, the information contained in the binary vector $\bm{z}$ at the output of the NPU is transmitted using pulse-based modulation with on-off keying (OOK). 

\subsection{Channel} We model the effect of the wireless channel with a   binary symmetric channel (BSC), accounting for the joint effects of the channel and the demodulation. Therefore, the conditional probability of the received sample   $\hat{z}_{i, t}$ given the corresponding input $  z_{i, t}$ is 
\begin{align}
p_{\mathrm{BSC}} \left ( \hat{z}_{i, t} | z_{i, t}\right )= 
\begin{cases}
1-\epsilon,  & \mathrm{if}~ \hat{z}_{i, t}=z_{i, t}, \\
\epsilon, & \mathrm{if}~ \hat{z}_{i, t}\neq z_{i,t},
\label{eq:BSC} 
\end{cases} 
\end{align}
where $\epsilon$ is the crossover, or bit flip,  probability. In practice, one can relate the crossover probability to the signal to noise ratio (SNR) per bit, $E_b/N_0$, of an additive Gaussian channel  as \begin{equation} \epsilon= Q\bigg(\frac{2E_b}{N_0}\bigg).
\label{eq:Q_function}\end{equation}We refer to \cite{chen2022neuromorphic,chen2023neuromorphic} for an analysis of more realistic channels. 

Using the channel model \eqref{eq:BSC} in the probability distribution \eqref{eq:SNN_model}, the conditional distribution of the received spiking signals $ \hat{\bm{z}}$ given the sensed signals  ${\bm{x}}$ is obtained by marginalizing over the encoded spikes ${\bm{z}}$ as 
\begin{equation} 
p_{\bm{\phi}} \left ( \hat{\bm{z}}\parallel \bm{x} \right ) = {\sum_{\bm{z}} p_{\bm{\phi}}\left ( \bm{z} \parallel \bm{x} \right )p_{\mathrm{BSC}}\left ( \hat{\bm{z}}|\bm{z} \right )}.
\end{equation}
Therefore, for each time $t$, we have the conditional probability \cite{choi2019neural}
\begin{align} 
    \begin{split}
p_{\bm{\phi}}\left({\bm{\hat{z}}_{t}} | \bm{{x}}^t\right ) = \prod_{i \in \mathcal{V}} & \left ( \sigma \left ( u_{i,t} \right )-2 \sigma \left ( u_{i,t} \right )\epsilon +\epsilon  \right )^{\hat{z}_{i,t}} \\
&\cdot \left ( 1-\sigma \left ( u_{i,t} \right ) +2\sigma \left ( u_{i,t} \right ) \epsilon -\epsilon \right ) ^ {\left (  1-\hat{z}_{i,t}\right )},  
    \end{split}   
    \label{q_noise}
\end{align}
where $\sigma \left ( u_{i,t} \right )$ represents the spiking probability of read-out neuron $i$.

\subsection{Receiver}  At the receiver side, i.e., at the edge server, an inference procedure is run to process the received feature vector $\hat{\bm{z}}$,  corrupted by the communication channel, producing the inference result $\hat{\bm{y}}$. To address the inference problem, the edge server can use conventional deep learning models. Here, we adopt a feedforward ANN model that incorporates batch processing with a buffer storing data as needed. 
 

\section{Semantic-Aware Communication via Directed Information Bottleneck}
\label{sec:S-VDIB}


In this section, we introduce the proposed design criterion for the optimization of the SNN at the transmitter. As explained in Sec. I, we follow the principle of providing a general design framework for the encoder, leaving open the possibility to design different decoders using conventional deep learning methods. The criterion follows the information
bottleneck  (IB) principle with the aim of
producing a spike-based compressed representation of the input signal that retains only the most relevant information for transmission on the wireless channel.

\subsection{Directed Information Bottleneck}

In order to account for the temporal nature of the signals being sensed and communicated, we resort to the \textit{directed} information bottleneck formulation (DIB) introduced in \cite{skatchkovsky2021learning}. In the DIB framework, the design goal is to maximize the directed information between the encoded signal  $\hat{\bm{Z}}$ and the target signal $\bm{Y}$, thus optimizing the inference performance, while minimizing the directed information between the sensed signal $\bm{X}$ and the encoded signal  $\hat{\bm{Z}}$ so as  to reduce the communication overhead by preserving only task-relevant information \cite{jang2019introduction}, \cite{skatchkovsky2021learning}. 

Accordingly, the design criterion is to  minimize the objective function 
\begin{equation}
\mathcal{L}_{DIB} ( \bm{\phi}  ) = - I  ( \hat{\bm{Z}} \to  \bm{Y}  ) + \beta \cdot I ( \bm{X} \to \hat{\bm{Z}} ) , \label{DIB}
\end{equation}
where $\beta > 0$ is a hyperparameter controlling the trade-off between the two terms,  and we recall that vector $\phi$ encompasses the parameters of the SNN. Note again that this criterion does not target the ANN at the receiver.  

\subsection{Variational Directed Information Bottleneck Reformulation}
In this section, we describe a practical formulation that allows optimization of the SNN parameters $\phi$. The presentation follows \cite{skatchkovsky2021learning}.

To evaluate the objective function $\mathcal{L}_{DIB}(\bm{\phi})$ in \eqref{DIB}, we start by characterizing the first directed information term, which can be decomposed as
\begin{align}
I(\hat{\bm{Z}}\to\bm{Y}) &=
H\left ( \bm{Y} \right ) - H ( \bm{Y}\parallel \hat{\bm{Z}} ) \\
&= H\left ( \bm{Y} \right ) + E_{p\left ( \bm{y},\hat{\bm{z}} \right )}\left [ \log p \left ( \bm{y}\parallel \hat{\bm{z}} \right ) \right ]. 
\end{align}
Similarly, for the second term we have 
\begin{align}
I( \bm{X} \to \hat{\bm{Z}})  
& = {E}_{p\left({\hat{\bm{z}}}, {\bm{x}}\right)}\left[\log \frac{p_{\bm{\phi}}\left({\hat{\bm{z}}}\parallel \bm{x}\right)}{p\left(\hat{\bm{z}}\right)}\right] .
\end{align}

Following the standard variational inference approach (see, e.g., \cite{simeone2022machine}), we introduce a variational  causally conditional distribution $q_{\bm{\theta}}\left(\bm{y}\parallel\hat{\bm{z}}\right)$ as
\begin{equation}
q_{\bm{\theta}}\left(\bm{y}\parallel\hat{\bm{z}}\right)=\prod_{t=1}^{T}q_{\bm{\theta}}\left ( \bm{y}_{t}|\bm{y}^{t-1}, \hat{\bm{z}}^{t} \right ), 
\end{equation}
where $\bm{\theta}$ is a learnable parameter vector. 
By using a standard variational bound, we then have the bound
\begin{align}
I  ( \hat{\bm{Z}} \to \bm{Y} ) 
&\geq H\left ( \bm{Y} \right ) + E_{p\left ( \bm{y},\hat{\bm{z}} \right )}\left [ \log q_{\bm{\theta}} \left ( \bm{y}\parallel \hat{\bm{z}} \right ) \right ]\nonumber\\
&= H\left ( \bm{Y} \right ) + \sum_{t=1}^{T} E_{p\left ( \bm{y^{t}},\hat{\bm{z}}^{t} \right )}\left [ \log q_{\bm{\theta}}\left ( \bm{y}_{t}|\bm{y}^{t-1}, \hat{\bm{z}}^{t} \right ) \right ].
\label{eq:Z_Y}
\end{align}
Similarly, choose a  variational prior
\begin{equation}
q(\hat{\bm{z}})=\prod_{t=1}^T q\left(\hat{\bm{z}}_{t} \mid \hat{\bm{z}}^{t-1}\right)
\end{equation}
to approximate the marginal distribution $p\left(\hat{\bm{z}}\right)$, we obtain the following bound 
\begin{align}
I  ( \bm{X} \to \hat{\bm{Z}} )  
&\leq {E}_{p\left({\hat{\bm{z}}}, {\bm{x}}\right)}\left[\log \frac{p_{\bm{\phi}}\left({\hat{\bm{z}}}\parallel \bm{x}\right)}{q\left(\hat{\bm{z}}\right)}\right]\nonumber\\
&=
\sum_{t=1}^{T}{E}_{p\left({\hat{\bm{z}}}^{t}, {\bm{x}}^{t}\right)}\left[\log \frac{p_{\bm{\phi}}\left({\hat{\bm{z}}}_{t} \mid {\bm{x}}^{t}, \hat{\bm{z}}^{t-1}\right)}{q\left(\hat{\bm{z}}_{t} \mid \hat{\bm{z}}^{t-1}\right)}\right].
\label{eq:X_Z}
\end{align}
By combining \eqref{eq:Z_Y}, \eqref{eq:X_Z}, we recast the objective function
in \eqref{DIB} as

\begin{align} \label{eq:VDIB}
    \begin{split}
&   \mathcal{L}_{VDIB}\left( \bm{\phi},\bm{\theta} \right)  = \\
&   E_{p(\bm{y},\bm{x})}\left\{ E_{p_{\bm{\phi}}(\hat{\bm{z}} \parallel \bm{x})} \left [ -\log q_{\bm{\theta}} \left ( \bm{y}||\hat{\bm{z}} \right ) + 
 \beta  \log \frac{p_{\bm{\phi}} \left ( \hat{\bm{z}}||\bm{x} \right ) }{q(\hat{\bm{z}})}\right]\right\}.
   \end{split}
\end{align}
We refer to the  design approach targeting objective (\ref{eq:VDIB}) as the  \emph{semantic variational directed information bottleneck} (S-VDIB). Please note that the entropy term $H(\bm{Y})$ does not occur in the optimization objective as it is a constant related to the input data distribution.  

In the S-VDIB formulation, the term $-\log q_{\bm{\theta}} \left ( \bm{y}||\hat{\bm{z}} \right )$ represents the classification loss, while
$\beta  \log \frac{p_{\bm{\phi}} \left ( \hat{\bm{z}}||\bm{x} \right ) }{q(\hat{\bm{z}})}$ acts as a regularization term. 
For compactness, in the following we use  $\ell_{\bm{\theta}}(\hat{\bm{z}}, \bm{y})$ and  $\ell_{\bm{\phi}}(\hat{\bm{z}}, \bm{x}) $,  respectively, to denote these two terms.

\subsection{Optimization with Monte Carlo Gradient Estimation}
The proposed S-VDIB minimizes the objective   $\mathcal{L}_{VDIB}\left( \bm{\phi},\bm{\theta} \right)$ via Monte Carlo gradient estimation, which is performed separately on the encoder model  $p_{\bm{\phi}}$ and the inference model  $q_{\bm{\theta}}$ as they apply different learning rules.  
The gradient with respect to parameters $\bm{\theta}$ inside the expectation, \begin{equation} 
\nabla_{\bm{\theta}} \mathcal{L}_{\mathrm{VDIB}}\left( \bm{\phi},\bm{\theta}  \right)=\mathrm{E}_{p(\bm{x}, \bm{y})} \mathrm{E}_{p_{\bm{\phi}}(\hat{\bm{z}} \| \bm{x})}
\left [ \nabla_{\bm{\theta}} \ell_{\bm{\theta}}(\hat{\bm{z}}, \bm{y}) \right ],
\end{equation} can be computed using standard backpropagation. 
In contrast, since the expectation over $p_{\bm{\phi}}$ makes it difficult to obtain the unbiased gradient estimator of encoder parameters $\bm{\phi}$, we use a generic method score function estimator \cite{williams1992simple} 
\begin{align} 
    \begin{split}
&   \nabla_{\bm{\phi}} \mathcal{L}_{\mathrm{VDIB}}\left( \bm{\phi},\bm{\theta}  \right)  = \\
&   \mathrm{E}_{p(\bm{x}, \bm{y})} \mathrm{E}_{p_{\bm{\phi}}(\hat{\bm{z}} \| \bm{x})}\biggl[\Bigl(\ell_{\bm{\theta}}(\hat{\bm{z}}, \bm{y}) + \beta  \ell_{\bm{\phi}}\left(\hat{\bm{z}}, \bm{x}\right) \Bigr) \nabla_{\bm{\phi}} \log p_{\bm{\phi}}(\hat{\bm{z}} \| \bm{x}) \biggr].\nonumber
 \label{VDIB}
   \end{split}     
\end{align}
Using the chain rule, the gradient with respect to the parameters $\phi_{i} = \left\{ w_{j, i},w_{i},\gamma_{i} \right\} $ at time $t$ has the following form
\begin{align} 
 \nabla_{\phi_{i}} \log p_{\bm{\phi}}\left(\hat{z}_{i, t} \| u_{i, t}\right) &=
\nabla_{u_{i, t}} \log p_{\bm{\phi}}\left(\hat{z}_{i, t} \| u_{i, t}\right) \nabla_{\phi_{i}} u_{i, t}.
\end{align}
Following from \eqref{eq:membrane_potential} and \eqref{q_noise}, we can compute
\begin{align*} 
& \nabla_{u_{i, t}} \log p_{\bm{\phi}}\left(\hat{z}_{i, t} \| u_{i, t}\right)  = \\
&  \left[ \frac{\hat{z}_{i,t}}{\sigma\left(u_{i,t}\right)+\frac{\varepsilon}{1-2 \varepsilon}}  + \frac{1-\hat{z}_{i,t}}{\sigma\left(u_{i, t}\right)+\frac{\varepsilon-1}{1-2 \varepsilon}} \right] \sigma\left(u_{i, t}\right) (1-\sigma\left(u_{i, t}\right)),
\end{align*}
and
\begin{align*} 
&\nabla_{\gamma_{i}} u_{i, t}=1 , \nonumber \\
&\nabla_{w_{j,i}}  u_{i, t} = \bm{a}_{t} \ast \bm{z}_{j, \leq t} ,\\
&\nabla_{w_{i}}  u_{i, t} = \bm{b}_{t} \ast \bm{z}_{i, \leq t-1} .
\end{align*} 
Given a finite dataset $\left\{ \left ( x^{(n)},y^{(n)} \right )\right\}_{n=1}^{N}$, the above gradients involving expectation can be approximated using Monte Carlo sampling as
\begin{align} 
\nabla_{\bm{\theta}} \mathcal{L}_{\mathrm{VDIB}}\left( \bm{\phi},\bm{\theta}  \right)\simeq \frac{1}{N}\sum_{n=1}^{N}
 \nabla_{\bm{\theta}} \ell_{\bm{\theta}}(\hat{\bm{z}}^{(n)}, \bm{y}^{(n)}),
 \label{grad_dec}
\end{align}
\begin{align} 
   \begin{split}
\nabla_{\bm{\phi}} \mathcal{L}_{\mathrm{VDIB}}\left( \bm{\phi},\bm{\theta} \right)  \simeq &  \frac{1}{N}\sum_{n=1}^{N} \biggl[\left(\ell_{\bm{\theta}}(\hat{\bm{z}}^{(n)}, \bm{y}^{(n)})  + 
 \beta \ell_{\bm{\phi}}(\hat{\bm{z}}^{(n)}, \bm{x}^{(n)})\right)  \\ &
 \nabla_{\bm{\phi}} \log p_{\bm{\phi}}(\hat{\bm{z}}^{(n)} \| \bm{x}^{(n)}) \biggr].     
    \end{split} 
    \label{grad_enc}
\end{align}Algorithm 1 summarizes the procedure.


\SetKwInput{KwData}{Input}
\SetKwInput{KwResult}{Output}
\RestyleAlgo{ruled}
\begin{algorithm}[hbt!]
\caption{S-VDIB}\label{alg:two}
\KwData{Dataset $\left\{ X, Y\right\}$, channel noise level $\epsilon$, number of time steps $T$, learning rate $\eta$, hyperparameter $\beta$}  
{Initialize the parameters of encoder $p_{\bm{\phi}}$ and decoder $q_{\bm{\theta}}$} \\
\For{each epoch}{
  Select a sample $\left ( \bm{x}, \bm{y} \right )$
  
  \For{each time step $t = 1,2,...,T$}{

  Generate corrupted signal $\hat{\bm{z}}_t$  based on {\eqref{q_noise}}

  Compute encoder loss  $\ell_{\bm{\phi}}(\hat{\bm{z}}, \bm{x}) $
  
  Generate output $\hat{\bm{y}}_t$ from inference network $q_{\bm{\theta}}$
  }

  }
  
  Compute loss $\mathcal{L}_{VDIB}\left ( \bm{\phi}, \bm{\theta } \right )$ based on \eqref{VDIB}

  Update parameters $\bm{\phi}$, $\bm{\theta}$ based on \eqref{grad_dec}, \eqref{grad_enc}
  
\end{algorithm}

\section{Simulation Results}
\label{sec:results}

\subsection{Simulation Setup}

In this section, we present numerical results to compare the performance of the proposed design approach to several benchmarks.

\paragraph{Dataset}
We adopt the following standard data sets.
\begin{itemize}
\item N-MNIST is recorded using a DVS camera moved through three saccadic motions, with each saccade lasting approximately $100$ $\text{ms}$. It comprises $60.000$ training images and $10.000$ test images. Each image is a $34 \times 34$ pixel representation of a handwritten digit ranging from $0$ to $9$. 

\item MNIST-DVS is obtained with a DVS camera to capture the slow movement of digits displayed on an LCD monitor. Each record lasts $2-3$ $\text{s}$, and the image resolution is $128 \times 128$ pixels. The dataset is available in three scales, and for this study, we use ``scale $4$". It consists of $10.000$  images ($9.000$ samples for training and $1.000$ samples for testing). Additionally, each image is cropped to $34\times 34$ pixels,  and only events within the first $2$ seconds of the record are used for analysis.
\end{itemize}
Following \cite{skatchkovsky2020end}, we pre-process the neuromorphic datasets by splitting each record into $T$ time steps and performing event accumulation to form frame-based representations. 
Given that event-based data includes polarity information, representing changes in pixel luminosity, we configure the input layer with $34 \times 34 \times 2$ neurons to leverage this information.

\begin{table}[H]
\caption{\centering{Layout of the communication system}}
\label{model}
\centering
\begin{tabular}{|c|c|c|}
\hline
                         & Layer           & Dimension \\ \hline
\multirow{2}*{Encoder SNN} & Input           & 34 $\times$ 34 $\times$ 2  \\ \cline{2-3} 
                         & SNN            & $k$       \\ \hline
Channel                  & BSC           & $k$       \\ \hline
\multirow{3}*{Inference ANN} & Flatten        & $k \times T$     \\ \cline{2-3} 
                         & Dense + Relu    & 1024      \\ \cline{2-3} 
                         & Dense + Sigmoid & 10        \\ \hline
\end{tabular}
\end{table}

\paragraph{System Architecture}
As illustrated in Table \ref{model},  the SNN encoder is a single-layer network with $k$ read-out neurons at the output layer. The transmitter communicates using an IR module and at each time instant $t$ the outputs $k$ read-out neurons are transmitted using pulse-based modulation (OOK), thus requiring $k$ channel uses (c.u), i.e. a communication budget (communication overhead) of $k$ bits. 
The inference ANN employs a multi-layer perceptron (MLP) model composed of two fully connected layers with a ReLU activation function for the hidden layer and a sigmoid activation function for the last layer. 
The input to the MLP model is the collection of the signals $\hat{\bm{z}}_{t}$ over the $T$ time steps. While other design criteria are possible, we adopt as the ANN decoder the trained decoder $q_\theta$ as per Algorithm 1. Following \cite{skatchkovsky2021learning}, we select the prior $q(\hat{\bm{z}})$ as a fixed Bernoulli distribution in the form of $q(\hat{\bm{z}}) = \prod_{t=1}^{T} \textrm{Bern} \left ( \hat{\bm{z}}_{t}\mid 0.3 \right )$.

\paragraph{Training Details}
In the training process, we draw one sample in the Monte Carlo estimator ($n = 1$ in \eqref{grad_dec}-\eqref{grad_enc}) to generate a discrete vector $\hat{\bm{z}}$ from the conditional distribution $p_{\bm{\phi}} \left({\hat{\bm{z}}} \parallel {\bm{x}}\right )$, while in the test experiments, vector $\bm{z}$ and $\hat{\bm{z}}$ are sampled individually according to the encoder output membrane potential and the crossover probability of the BSC (i.e. the corresponding $E_b/N_o$ level). The hyperparameter $\beta$  is selected from the range $ \left[10^{-4}, 10^{-1}\right]$. We set $T = 50$ for MNIST-DVS and $T = 48$ for N-MNIST.

\paragraph{Baselines}
Following \cite{skatchkovsky2020end,chen2022neuromorphic}, we select two ANN-based conventional schemes as baselines, namely separate source-channel coding (SSCC) and joint source-channel coding (JSCC). 
The SSCC scheme consists of a Vector Quantized-Variational AutoEncoder (VQ-VAE) model \cite{van2017neural} in combination with a rate-1/2 LDPC code. 
The received data is processed by belief propagation by the VQ-VAE decoder, and then passed through an ANN classifier. Specifically, the reconstructed data across $T$ time steps are collected and fed into an ANN classier with a fully connected hidden layer of 1024 neurons. For the JSCC scheme, we choose the learning-based model from \cite{song2020infomax}, which maps input data directly to noisy channel outputs via a JSCC encoder, trained on our datasets.

\subsection{Results} 

\paragraph{Accuracy vs. SNR per Bit}

\begin{figure*}[t]
\centerline{\includegraphics[width=0.9\textwidth]{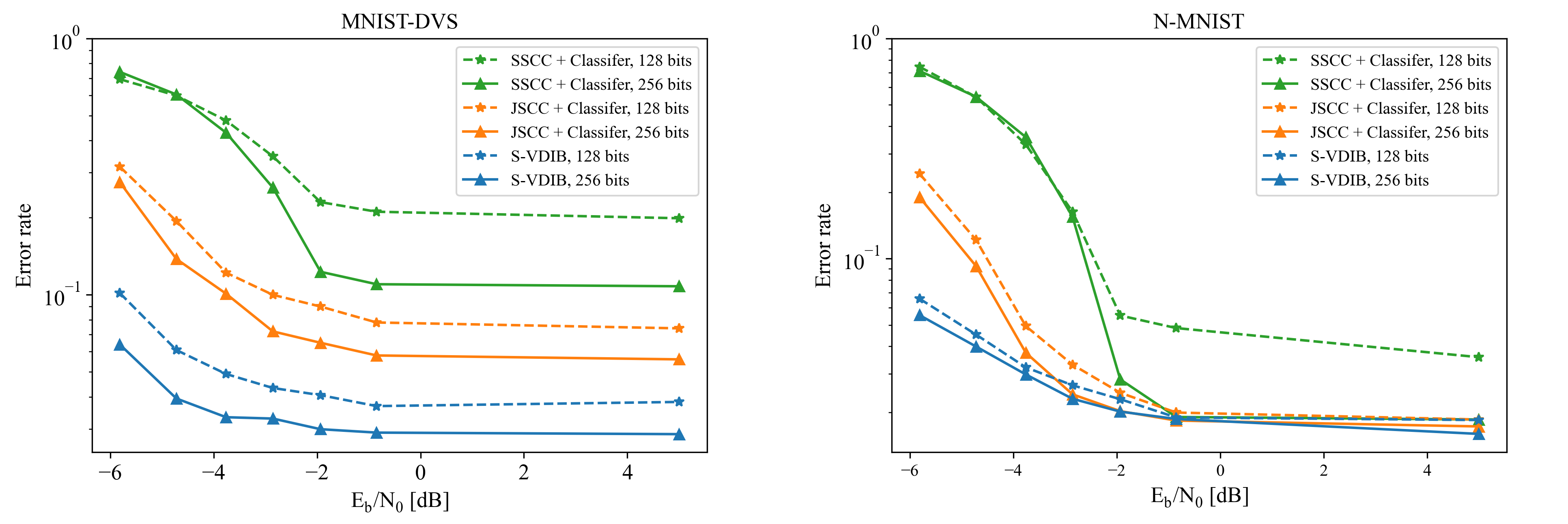}}
\caption{Test error rate as a function of the SNR per bit, $E_b / N_0$,  for the two standard datasets MNIST-DVS (left) and N-MNIST (right).    
}
\label{fig:error_rate_main}
\end{figure*}
We start by fixing the number of channel uses for communication, which coincides with the number of neurons $k$ at the encoder output layer. Fig.~\ref{fig:error_rate_main} depicts the classification {error rate} as a function of the SNR per bit, $E_b/N_0$, in (\ref{eq:Q_function}) for $k = 128$ and $k=256$. We observe that S-VDIB outperforms the two baseline methods across nearly all SNR levels, with the difference in performance being more pronounced in the low-SNR regime. In this regard, we note that, for the N-MNIST dataset, JSCC achieves  performance levels competitive with S-VDIB under sufficiently larger SNR, but its accuracy experiences a sharp decline at around $E_b / N_0 \leq -3 $ dB.  
These results suggest that, under a constrained communication budget, task-oriented communication can effectively extract and transmit robust features in comparison to conventional communication schemes, thereby improving accuracy.

\paragraph{Robustness against a Train-Test SNR Mismatch}
\begin{figure}[h!]
\centerline{\includegraphics[width=0.4\textwidth]{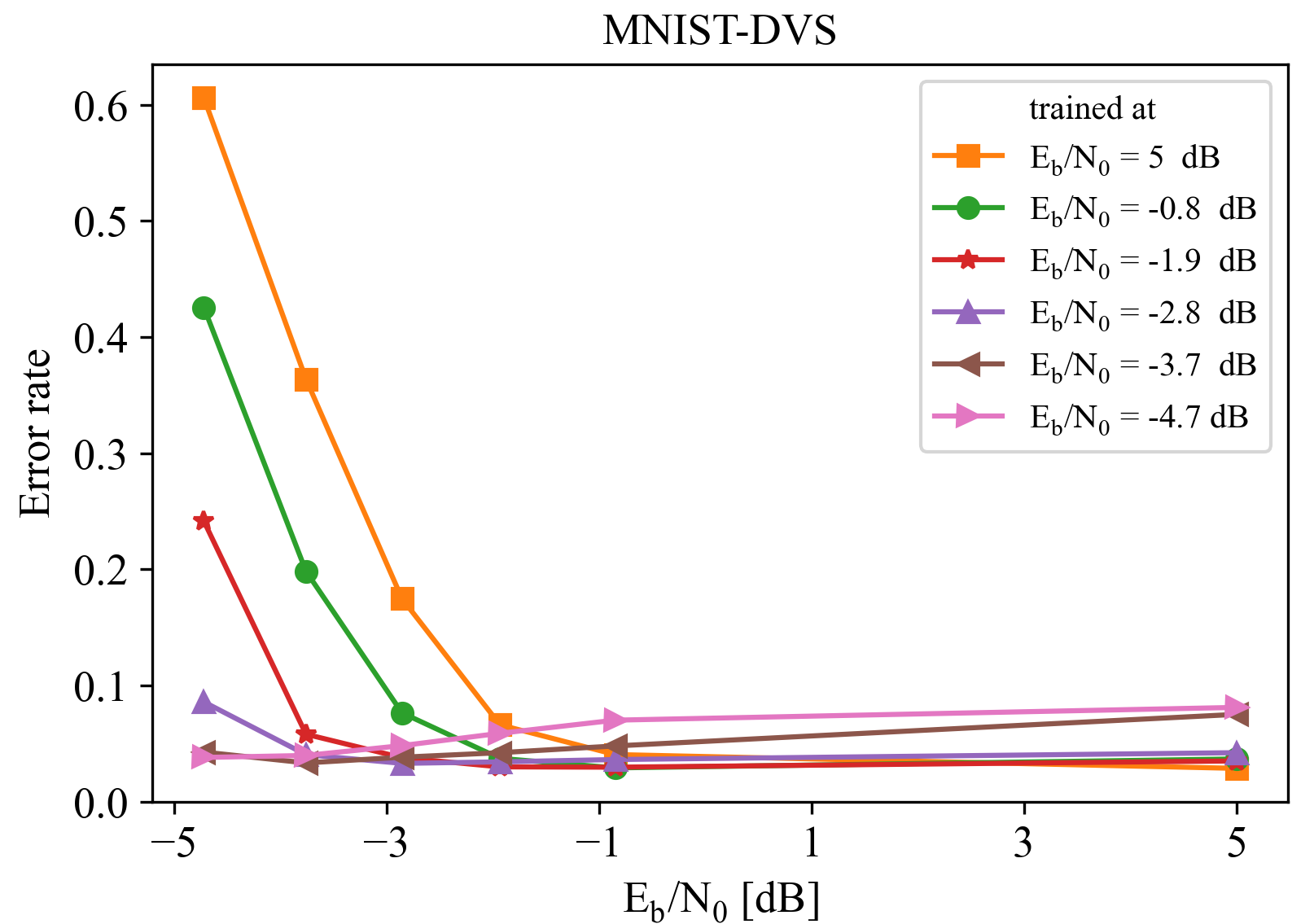}}
\caption{Test error rate for S-VDIB on the MNIST-DVS dataset with an SNR mismatch between training and testing conditions ($k=256$). }
\label{fig:error_rate_channel_missmatch}
\end{figure}
We now evaluate the robustness of S-VDIB when faced with a mismatch between training and test SNR conditions. Specifically, we train the model at a given level of SNR per bit, and vary the SNR experienced at test time. 
As depicted in Fig.~\ref{fig:error_rate_channel_missmatch}, the model originally trained at a moderate noise level 
suffers only minor degradation at other noise levels, demonstrating a capacity to maintain a reasonably high performance level even in the presence of an SNR mismatch.

\paragraph{Trade-off between Accuracy and Sparsity}
\begin{figure}[t]
\centerline{\includegraphics[width=0.42\textwidth]{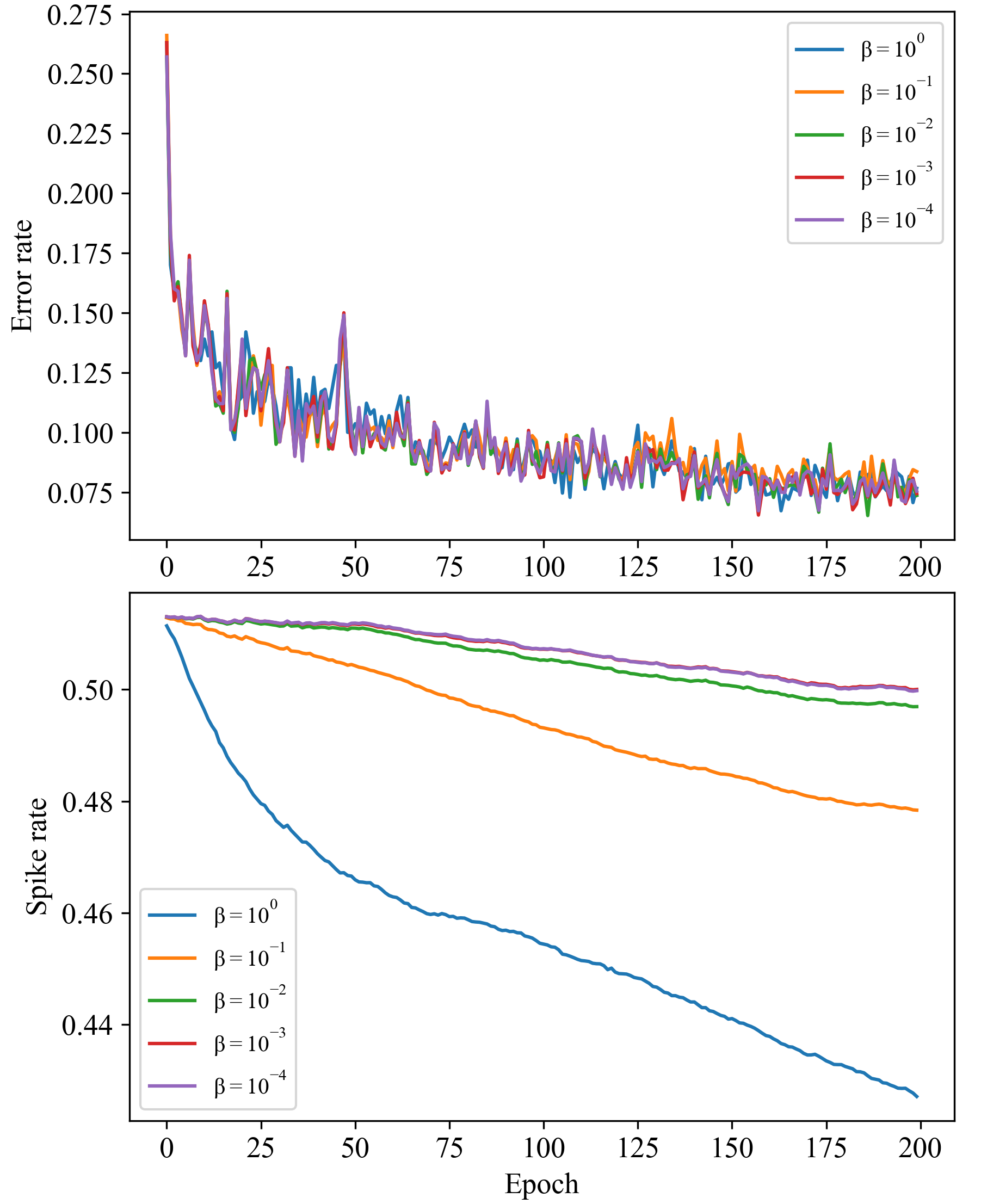}}
\caption{(top) Test classification error and (bottom) per-neuron spike rate as a function of the training epoch number for the MNIST-DVS dataset  ($k=256$, $E_b/N_0=-4.7$ dB). 
}
\label{fig:accuracy_sparsity}
\end{figure}
In the goal-oriented communication system at hand, a rate-distortion trade-off may be formulated as the trade-off between sparsity, quantified by the average spike rate of the encoder's output, and the classification accuracy. 
Here, we explore this trade-off by varying the hyperparameter  $\beta$ in the DIB design criterion (\ref{DIB}). 
Specifically, Fig.~\ref{fig:accuracy_sparsity} shows that the test error and the spike rate per neuron as a function of the training epoch for different values of the hyperparameter $\beta$.

It is observed that an increase in the value of hyperparameter $\beta$  leads to a progressively sparser encoded output, entailing a lower energy consumption. Furthermore, within the considered range of values of $\beta \in \left[10^{-4}, 10^{0}\right]$, sparsity is decreased with a larger $\beta$ with only a negligible drop in accuracy. Further simulations, not reported here, demonstrate that for larger values of $\beta$, e.~g., $\beta \approx 10^{2}$, the information contained in the input is overly compressed, resulting in a significant decrease in classification accuracy.

\section{Preliminary Testbed}
\label{sec:testbed}
\begin{figure}[ht]
\centerline{\includegraphics[scale=0.338]{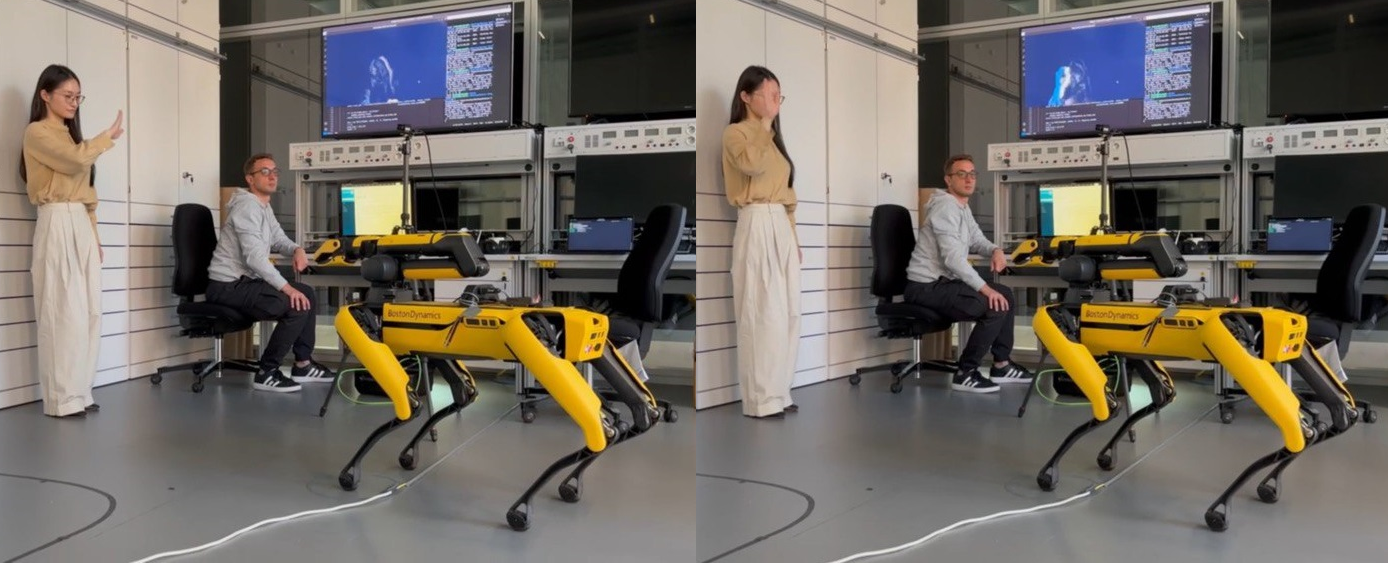}}
\caption{Photo of the current lab setup. 
} 
\label{fig:lab_setup}
\end{figure}

This section reports on ongoing work to set up a testbed to validate the principle of neuromorphic communication introduced in \cite{skatchkovsky2020end,chen2022neuromorphic} and further studied in this paper for the specific setting of device-edge co-inference.  As illustrated in Fig.~\ref{fig:lab_setup},  the testbed implements a wireless robotic control application based on gesture recognition via a neuromorphic sensor. 

\begin{figure}[h]
\centerline{\includegraphics[width=0.45\textwidth]{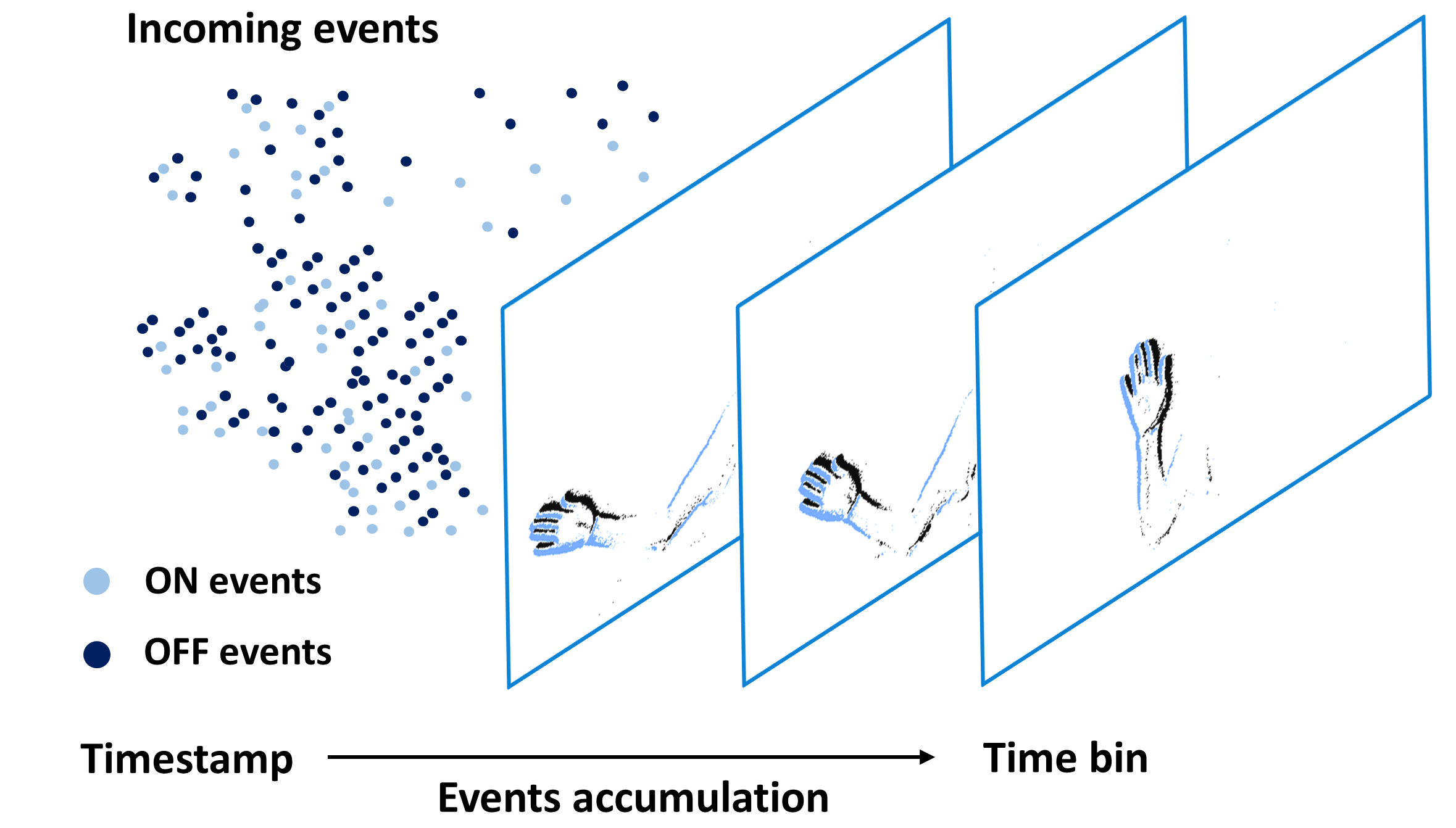}}
\caption{Data pre-processing for the gesture recognition task: events within a time bin are downsampled and fed to an SNN for further processing.}
\label{fig:data_preprocessing}
\end{figure}

Specifically, a Prophesee event-based camera is used as a neuromorphic sensor to capture  gestures of a user \cite{Proph}. The captured events are fed to a pre-preprocessing module,  where, as depicted in  Fig.~\ref{fig:data_preprocessing}, they are converted to spiking signals to be input  to the encoding  SNN. 

In the current implementation, the SNN performs only feature extraction by learning how to extract the task-relevant features from the processed camera inputs via the framework introduced in this paper.  
Based on the inference result, a control signal is transmitted to the robot, namely  Spot\copyright~\cite{Boston}, to replicate the gesture of the human. The current lab setup is shown in Fig.~\ref{fig:lab_setup}.

In a concurrent implementation, which is currently under development, the SNN will perform both feature extraction and channel coding by learning how to map the input signals directly to the channel input symbols. The system will operate over a wireless communication channel with impulse-radio transmission/reception.


\section{Conclusions}
\label{sec:conclusions}

In this work, we introduced a new system solution for device-edge co-inference that targets energy efficiency at the end device using neuromorphic hardware and signal processing,   while implementing conventional radio and computing technologies at the edge. 
The investigated communication scheme combines on-device SNN and server-based ANN, leveraging variational directed information bottleneck technique to perform inference tasks. From a deployment perspective, our model demonstrates superior performance, less need for communication overhead and robustness under time-varying channel conditions, implying promising potential in the further 6G work. Aspects of the proposed system solution were validated in a preliminary testbed setup that implements a wireless robotic control application based on gesture recognition via a neuromorphic sensor. The testbed setup is currently being expanded to integrate end-to-end learning via an impulse-radio communication link. The proposed architecture and the corresponding testbed setup are general in the sense that they can support the implementation of different semantic tasks. Besides applications in robotics, in the future, we will also consider bio-medical applications that leverage the energy and communication efficiency of this architecture.

\bibliographystyle{ieeetr}
\bibliography{bibliography}

\clearpage

\end{document}